\documentclass[10pt,twocolumn,letterpaper]{article}

\usepackage{cvpr}              
\usepackage{stmaryrd}
\usepackage{tabularray}
\usepackage{graphicx}
\usepackage{booktabs}
\usepackage{multirow}
\usepackage{rotating}
\usepackage[normalem]{ulem}
\usepackage{enumitem}
\usepackage{graphicx}
\usepackage{subcaption}
\usepackage{caption}
\usepackage[margin=1in]{geometry}  
\usepackage{float}
\usepackage{dashrule}

\definecolor{cvprblue}{rgb}{0.21,0.49,0.74}
\usepackage[pagebackref,breaklinks,colorlinks,allcolors=cvprblue]{hyperref}

\def\paperID{8039} 
\def\confName{CVPR}
\def\confYear{2025}

\title{Joint Optimization of Neural Radiance Fields and Continuous Camera Motion from a Monocular Video}

\author{
\text{Hoang Chuong Nguyen}\textsuperscript{1} \hspace{0.5cm} \text{Wei Mao} \hspace{0.5cm}  \text{Jose M. Alvarez}\textsuperscript{2}  \hspace{0.5cm}
\text{Miaomiao Liu}\textsuperscript{1}\\
\textsuperscript{1}\text{Australian National University} \hspace{1.0cm}  \textsuperscript{2}\text{NVIDIA}\\
{\tt\small {\{hoangchuong.nguyen, miaomiao.liu\}@anu.edu.au } \hspace{0.5cm} josea@nvidia.com}
}

\begin{document}
\twocolumn[{%
\renewcommand\twocolumn[1][]{#1}%
\maketitle
\begin{center}
    \captionsetup{type=figure}
    \begin{minipage}{0.495\textwidth}
        \centering
        \includegraphics[width=\linewidth]{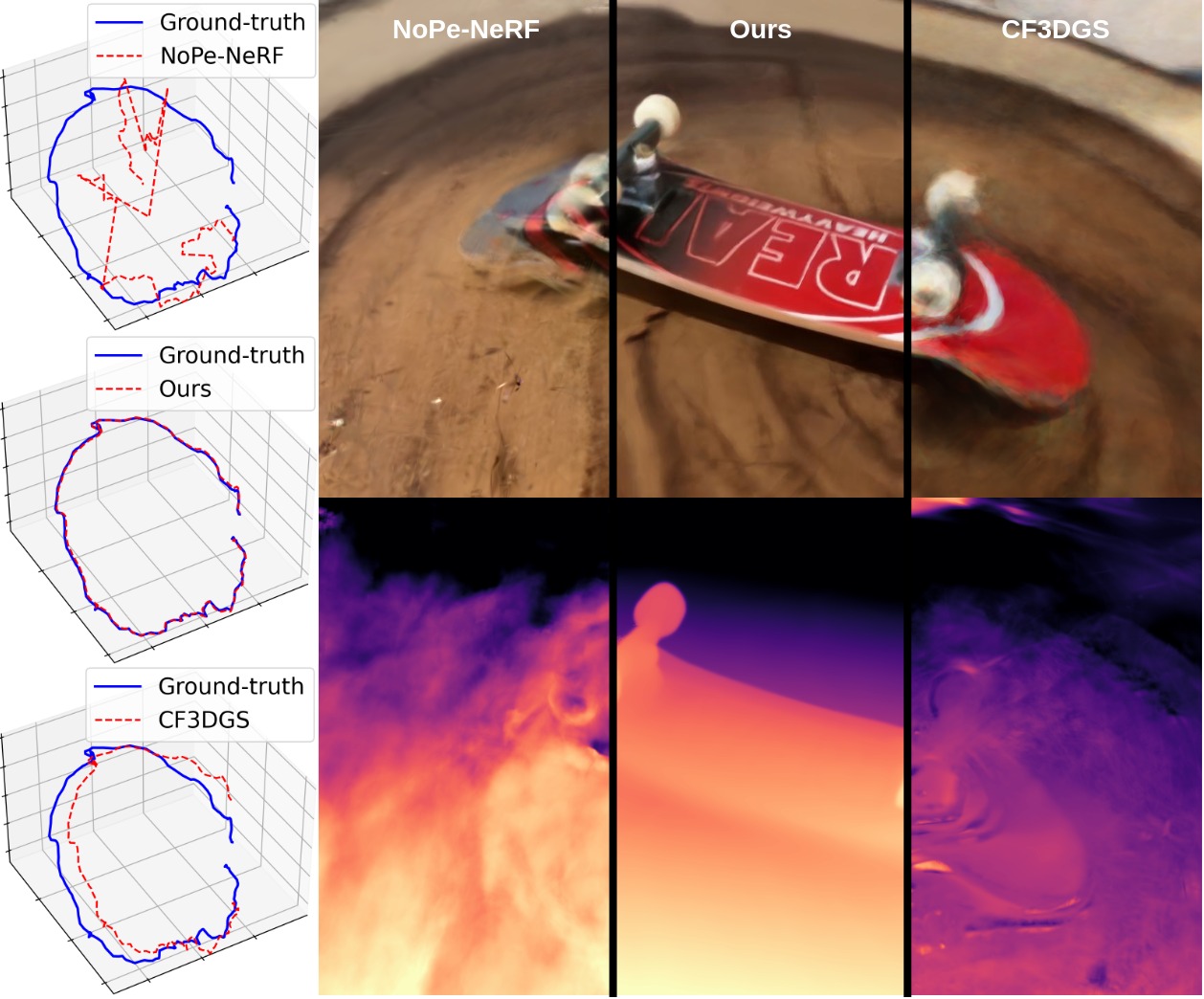}
    \end{minipage}
    \hfill
    \begin{minipage}{0.495\textwidth}
        \centering
        \includegraphics[width=\linewidth]{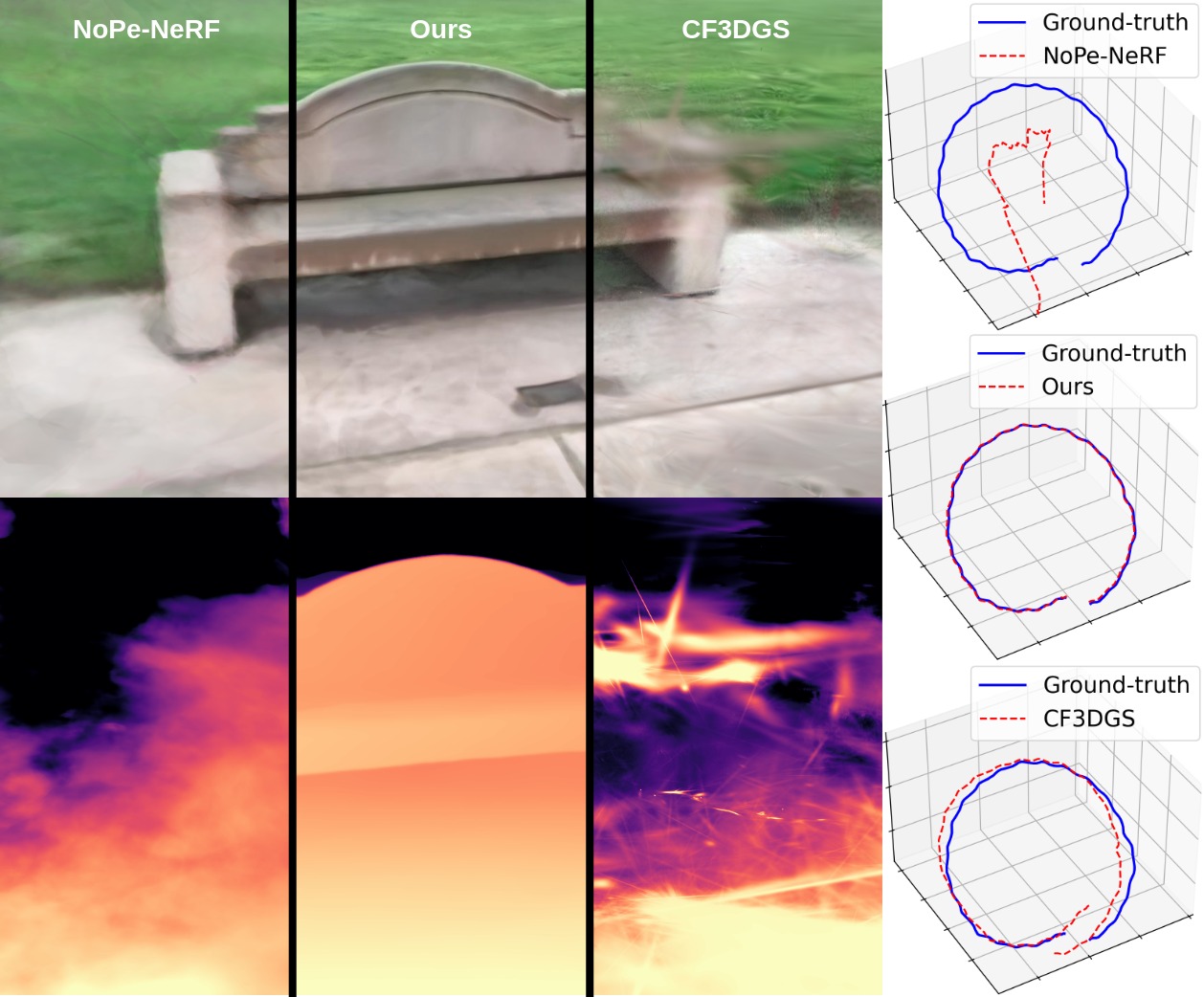}
    \end{minipage}
    \caption{\textbf{Comparison with previous works} \cite{bian2023nope, fu2023colmapfree}. Our method achieves superior performance in camera pose estimation (side), depth estimation (bottom) and novel-view synthesis (top).}\label{fig:intro_figure}
\end{center}%
}]

\begin{abstract}
Neural Radiance Fields (NeRF) has demonstrated its superior capability to represent 3D geometry but require accurately precomputed camera poses during training. To mitigate this requirement, existing methods jointly optimize camera poses and NeRF often relying on good pose initialisation or depth priors. However, these approaches struggle in challenging scenarios, such as large rotations, as they map each camera to a world coordinate system.
We propose a novel method that eliminates prior dependencies by modeling continuous camera motions as time-dependent angular velocity and velocity. Relative motions between cameras are learned first via velocity integration, while camera poses can be obtained by aggregating such relative motions up to a world coordinate system defined at a single time step within the video. Specifically, accurate continuous camera movements are learned through a time-dependent NeRF, which captures local scene geometry and motion by training from neighboring frames for each time step. The learned motions enable fine-tuning the NeRF to represent the full scene geometry.
Experiments on Co3D and Scannet show our approach achieves superior camera pose and depth estimation and comparable novel-view synthesis performance compared to state-of-the-art methods. Our code is available at \small \url{https://github.com/HoangChuongNguyen/cope-nerf}. 
\end{abstract}

\section{Introduction}
\label{sec:introduction}
Recent advances in Neural Radiance Fields (NeRF) \cite{mildenhall2021nerf, wang2021neus, yariv2021volume} have demonstrated impressive performance in synthesizing photorealistic images and representing 3D scenes. The training process of these models requires known camera poses for training images, which are usually obtained using Structure-from-Motion methods such as COLMAP \cite{schonberger2016structure}. However, this process introduces additional computation, lacks differentiability, and is sensitive to hand-crafted features \cite{lowe2004distinctive, bay2006surf}.

To remove its reliance on COLMAP, existing methods have attempted to jointly optimize NeRF and camera poses~\cite{wang2021nerf,lin2021barf,bian2023nope}. Despite their promising results, they either are limited to forward-facing scenes~\cite{wang2021nerf} or rely on prior knowledge such as a good camera pose initialization~\cite{lin2021barf}, or monocular depth~\cite{bian2023nope}. For the prior-based methods, their performance highly depends on the priors, and due to their inherent strategy of learning the frame-wise camera-to-world mappings, they also struggle to handle large camera motions. 

In this paper, we propose a method to jointly optimize NeRF and camera poses which better leverages the temporal information of a monocular video~\emph{without relying on priors}. Specifically, instead of optimizing the discrete camera pose for each frame, we predict a continuous camera motion represented as angular velocity and velocity, capturing the temporal continuity and smoothness of the camera movements. Such continuous representation breaks camera motion down into infinitely small intervals, enabling large movements between any camera and world frame to be computed through the aggregation of motion over time via velocity integration. This approach significantly simplifies learning large camera motions, as shown in Fig.~\ref{fig:intro_figure}.

Given camera pose for each frame obtained through the velocity integration, one can train a single NeRF defined in a global world coordinate system, similar to existing approaches. 
However, camera poses are often noisy during the early training stage. Mapping 3D points from camera to the world coordinate system using these noisy poses makes the training harder and often stuck in local minimum. To address this, we propose to adopt a time-dependent NeRF to represent the scene at each time step, trained using neighboring frames around time $t$. While the time-dependent NeRF at each time step can only capture local scene geometry, it is sufficient for estimating continuous camera motion at the corresponding time step. Additionally, our modeling allows us to constrain the continuous camera motion and the time-dependent scene geometry represented by Signed Distance Field (SDF) with the linear relationship between the SDF flow and scene flow to improve the camera motion estimation as introduced in~\cite{maoneural}.

To model the full scene geometry, we fix the camera motion in the later training stage and continue training the time-dependent NeRF corresponding to the world coordinate system. Thanks to our continuous camera motion modeling, we can integrate the motion of each camera frame relative to the world coordinate system and train the NeRF using all available frames. This process enables us to reconstruct the entire scene geometry while improving rendering quality. Our proposed framework allows us to obtain an accurate camera pose for each frame as well as the 3D scene representation. In summary, our contributions are as follows:
\begin{itemize}
    \item We introduce a prior-free pipeline to jointly optimize camera motion and scene geometry using NeRF.
    \item We propose to represent the camera motion in a continuous way, helping to learn large camera motions.
    \item We further propose to utilize a time-dependent NeRF, which allows the estimation of accurate camera motion within a small time window. It also enables the introduction of the linear relationship between the camera motion and scene geometry during training for further improving the camera motion estimation.
\end{itemize}
Our approach outperforms state-of-the-art methods in camera pose and depth estimation, and achieve comparable performance in novel-view synthesis.
  
\section{Related Works}

\noindent\textbf{Scene reconstruction and rendering with known camera poses.} 
Neural radiance fields (NeRF) \cite{mildenhall2021nerf}, and 3D Gaussian splatting (3DGS) \cite{kerbl3Dgaussians} have drawn attention in recent years due to their superiority and efficiency in rendering photo-realistic images and modeling 3D scenes. 
Extracting accurate 3D geometry with 3DGS remains an active research direction, with recent works introducing better rasterization strategies \cite{yu2024gaussian, zhang2024rade} and additional regularizations \cite{guedon2024sugar, huang20242d}. Regarding NeRF, several works have proposed to enhance neural implicit representations via SDF field \cite{wang2021neus, yariv2021volume}. Additional efforts have been made for accelerating the rendering process \cite{chen2022tensorf, muller2022instant}, 
reconstructing large scenes \cite{tancik2022block} or dynamic scenes \cite{maoneural, gao2021dynamic}. 
Despite their advancements, both these 3DGS-based and NeRF-based methods still rely on Structure-from-Motion (SfM) algorithms \cite{schonberger2016structure} to acquire the precomputed camera poses essential for training. Unlike these works, our proposed method can jointly optimize the camera poses and the scene geometry (represented via NeRF).

\noindent\textbf{Joint 3DGS and poses optimization.} Several works \cite{fu2023colmapfree, Matsuki:Murai:etal:CVPR2024, yan2024gs, sandstrom2024splat} have explored the joint optimization of 3DGS and camera poses. These methods typically employ a SLAM-style pipeline to sequentially optimize relative camera transformation between consecutive frames in a video. Contrary to these methods, we differ by modeling the continuous camera motions instead of the relative camera motions.
Such continuous motions enable us to enforce a constraint on the temporal changes of the observed scene geometry, thereby improving both the inferred camera poses and overall scene geometry.
However, this cannot be achieved by learning only the relative camera mappings. Additionally, without geometric prior, 3DGS-based methods have demonstrated limited success in scene reconstruction \cite{kerbl3Dgaussians, Matsuki:Murai:etal:CVPR2024}. The key challenge of this stems from their Gaussians densification which can be challenging near regions with abrupt depth change. Moreover, many noisy Gaussians can be grown when the poses are not fully optimized, potentially leading to memory issues during training \cite{fu2023colmapfree, kheradmand20243d}. Additionally, the performance of 3DGS-based methods tends to degrade when accurate 3D points for initializing the Gaussians are not available \cite{Matsuki:Murai:etal:CVPR2024, kerbl3Dgaussians}. To overcome this, either ground-truth depth \cite{yan2024gs, keetha2024splatam, Matsuki:Murai:etal:CVPR2024} or a pre-trained depth network \cite{fu2023colmapfree, sandstrom2024splat} is essential for initialization and guiding the densification process. In contrast, our method  does not rely on additional geometric priors and still achieves accurate 3D reconstruction.

\noindent\textbf{Joint NeRF and poses optimization.} 
Differing from 3DGS, NeRF-based methods do not require accurate point clouds for initialization, yet still outperform 3DGS-based approaches in reconstructing the scene geometry \cite{yu2024gaussian}. 
Regarding the joint NeRF and poses optimization, a pioneering method in this direction is i-NeRF \cite{yeninerf}, which utilizes a pre-trained NeRF model to recover camera poses. NeRFmm \cite{wang2021nerf} further advances this by jointly optimizing NeRF along with camera extrinsic and intrinsic parameters. BARF \cite{lin2021barf} and GARF \cite{chng2022gaussian} focus on addressing the negative impact of naive positional embeddings on pose estimation. While these methods have shown promising results, they are either limited to forward-facing scenes or require good pose initialization. NoPe-NeRF \cite{bian2023nope} tackles this by leveraging a pre-trained depth network \cite{ranftl2021vision}. However, it still struggles in dealing with large camera-to-world mappings. Unlike previous works, our method is prior-free and capable of handling large motions by exploiting temporal information from a monocular video.

In particular, we propose to model continuous camera motion and a time-dependent NeRF, which enables the learning of accurate short-term camera motions. As opposed to prior methods, our approach can avoid the direct optimization of large camera-to-world mappings, since the camera pose at any frame can be recovered by integrating the learned continuous camera motion up to a selected world frame. While the time-dependent NeRF is also leveraged in \cite{gao2021dynamic}, it is used for motion predictions in a dynamic scene. On the other hand, in our pipeline, this model offers an efficient way to compute the SDF flows, which is important to constrain the learning of the continuous camera motion. Compared to \cite{maoneural} which integrates the SDF flows overtime to obtain SDF values, our approach in leveraging this information is more efficient, as we can avoid the time-consuming SDF flows integration by directly predicting the SDF and deriving the SDF flows through automatic differentiation \cite{paszke2017automatic}.

\begin{figure*}
    \centering
    \includegraphics[width=\textwidth]{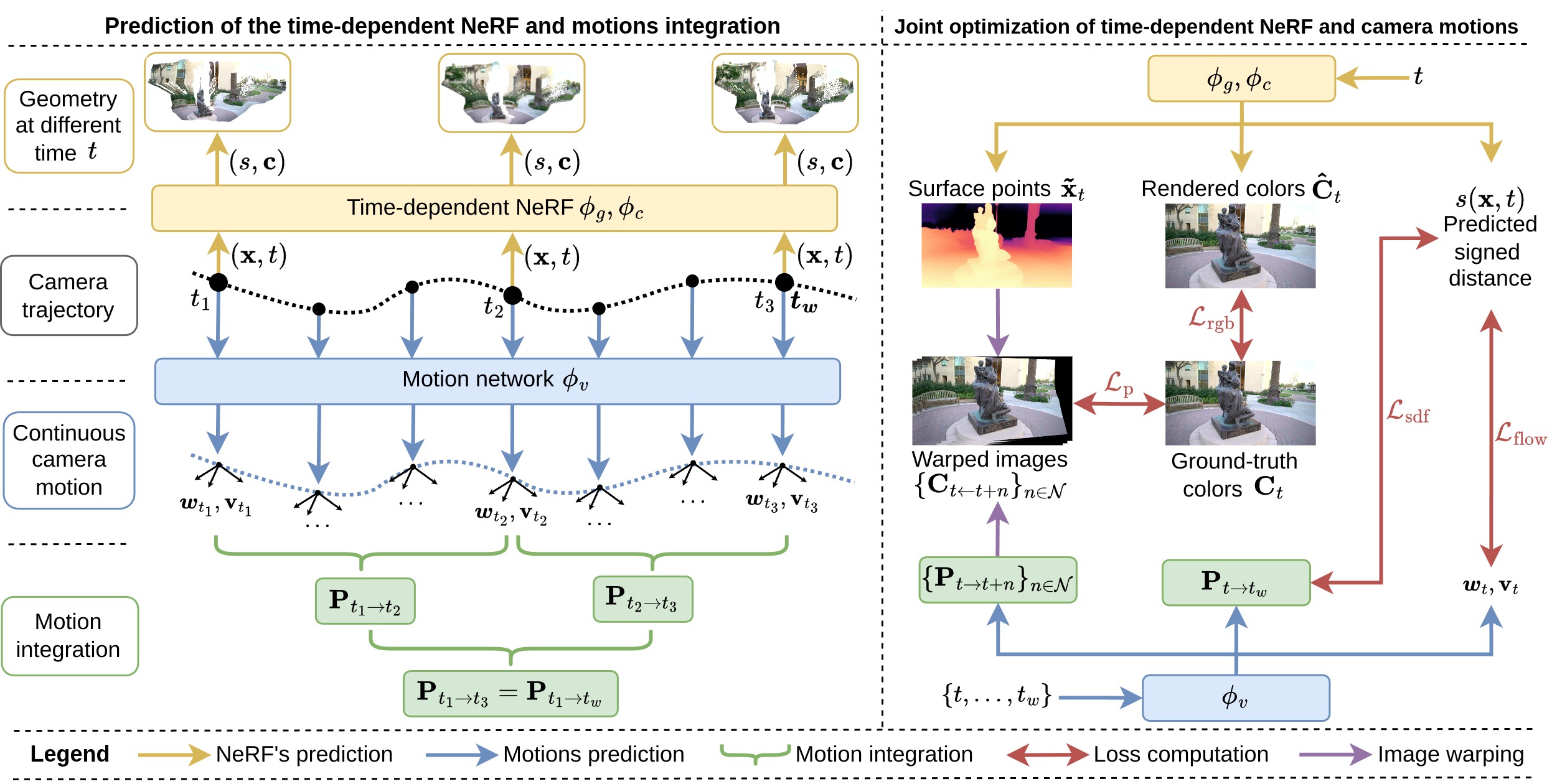}
    \caption{\textbf{Overview of our method}. \textbf{Left: } We jointly optimize the camera motion network $\phi_{v}$ to obtain the continuous camera motion represented as angular velocity $\boldsymbol{\omega}$ and velocity $\mathbf{v}$, and a time-dependent NeRF $(\phi_{g}\;,\phi_{c})$ to obtain the scene geometry (represented as Signed Distance Field) and appearance at different time steps. The camera velocities can be integrated to obtain the camera transformation $\mathbf{P}$ between any two frames. Such relative camera motion is used to get 3D correspondences across different time steps. \textbf{Right: } The training our pipeline involves the standard rendering loss $\mathcal{L}_{\text{rgb}}$, and several consistency losses including, the consistency between SDF and camera motion $\mathcal{L}_{\text{flow}}$, the photometric consistency $\mathcal{L}_{\text{photo}}$, and the geometry consistency loss $\mathcal{L}_{\text{sdf}}$.
    }
    \label{fig:method_overview}
\end{figure*}

\section{Method}
We tackle the problem of jointly optimizing camera motion and NeRF from a monocular video. Our goal is not limited to synthesizing photo-realistic images but also to achieving accurate camera motion and scene geometry estimation. We assume no additional information beyond camera intrinsics and images from the input video. 

Our method begins with the joint optimization of continuous camera motions (Sec.~\ref{subsec:joint_geometric_poses}) and a time-dependent NeRF (Sec.~\ref{subsec:time_dependent_nerf}). ~Fig.~\ref{fig:method_overview} depicts the overall pipeline of this joint optimization. Subsequently, camera poses are derived from the learned motions, which are used to fine-tune our previously trained model following the conventional NeRF training pipeline~\cite{wang2021neus} (Sec.~\ref{subsec:overall_training_pipeline}). 

\subsection{Preliminary: NeRF with SDF Presentation }
\label{subsec:preliminary}
In this work, we adopt the SDF-based NeRF model introduced in \cite{wang2021neus}. This method represents the scene appearance using a mapping function, $\mathbf{c} = \phi_c(\mathbf{x},\mathbf{d})$, that takes a point $\mathbf{x}\in\mathbb{R}^3$ and a camera viewing direction $\mathbf{d}\in \mathbb{R}^2$ as inputs to predict color $\mathbf{c}\in\mathbb{R}^3$. Likewise, the scene geometry is modeled via $s = \phi_g(\mathbf{x})$ which maps a 3D point $\mathbf{x}$ to a signed distance value $s \in \mathbb{R}$. For each camera ray $\mathbf{r}(h)=\mathbf{o}+h\mathbf{d}$ with viewing direction $\mathbf{d}$ passing through a camera origin $\mathbf{o} \in \mathbb{R}^3$, a set of $K$ 3D points can be sampled: $\{ \mathbf{x}_i (\mathbf{r})= \mathbf{r}(h_i) \mid i = 1, \ldots, K, \, h_i < h_{i+1} \}$. The signed distance can be converted into density value $\sigma$ using:
\begin{align}
    \sigma_i &= \max \left( \frac{\Phi \left( s_i \right) - \Phi \left( s_{i+1} \right)}{\Phi \left( s_i \right)}, 0 \right)\;, 
\end{align}
where  $\Phi(s) = \frac{1}{1 + e^{-\gamma s}}$, and $\gamma\in \mathbb{R}$ is a learnable parameter and $i$ denotes an index of a sampled point. The pixel's color can then be obtained via volume rendering:
\begin{equation}
    \mathbf{\hat{C}}({\bf r}) = \sum_{i=1}^{K} L_i \sigma_i \mathbf{c}_i\;,\label{eq:volume_rendering}
\end{equation}
with $L_i = \prod_{j=1}^{i-1} (1 - \sigma_j)$ being the accumulated transmittance along a ray.

\subsection{Continuous Camera Motion}
\label{subsec:joint_geometric_poses}
Previous methods \cite{bian2023nope, wang2021nerf, lin2021barf} aim to jointly optimize independent camera poses and a NeRF model to explain the observed images $\{  \mathbf{C}_t  \in \mathbb{R}^{H\times W \times 3} \}_{t=1}^{T}$. 
However, these approaches struggle in scenarios with significant camera-to-world transformations even with monocular depth prior from a pre-trained model~\cite{bian2023nope}.
Instead of relying on any prior, we propose to model the continuous camera motion by leveraging the temporal information from a video sequence to estimate the camera motions within a small time interval, thereby avoiding the direct estimation of the camera-to-world transformations. 
Specifically, we model this continuous motions as angular velocities and velocities of the camera. An MLP $\phi_v$ is then used to predict these velocities as:
\begin{equation}
    \boldsymbol{\omega}(t), \mathbf{v}(t) = \phi_v(t)\;,
\end{equation}
with $\boldsymbol{\omega}(t), \mathbf{v}(t) \in \mathbb{R}^3$ being the angular velocity and the velocity of the camera at time $t$. The relationship between the camera transformation and such velocities can then be expressed as,
\begin{equation}
    \frac{d\mathbf{R}}{dt} = [\boldsymbol{\omega}(t)]_{\times}\mathbf{R}\;, \quad \quad \quad \frac{d\mathbf{t}}{dt} = \mathbf{v}(t)\;,
\end{equation}
where $\mathbf{R}\in SO(3)$ and $\mathbf{t}\in \mathbb{R}^3$ are the camera rotation and translation, respectively. $[\boldsymbol{\omega}(t)]_{\times}\in so(3)$ is the skew-symmetric matrix of $\boldsymbol{\omega}(t)$. In practice, we use Euler method to solve the above Ordinary Differential Equations. In particular, the relative rotation and translation from time $t$ to $t+l$ ($l>0$) can be computed as,
\begin{align}
    \mathbf{R}_{t\rightarrow t+l} &= \prod_{u=0}^{U-1} \psi(\boldsymbol{\omega}(t+u\Delta t)\Delta t) \label{eq:relative_rot} \\
    \mathbf{t}_{t\rightarrow t+l} &= \sum_{u=0}^{U-1} \mathbf{v}(t +u\Delta t)\Delta t\;,\label{eq:relative_trans}
\end{align}
where $\psi(\cdot):\mathbb{R}^3\rightarrow SO(3)$ converts rotation angles to rotation matrix and $\Delta t = \frac{l}{U}$ is the step size. Note that we apply right-matrix multiplication in Eq. \ref{eq:relative_rot}. The transformation matrix that transform the coordinate of any point from any time step $t_1$ to another time step $t_2$ can be defined as,
\begin{align}
    \mathbf{P}_{t_1\rightarrow t_2} = 
    \begin{cases} 
        \mathbf{B}_{t_1\rightarrow t_2} & \text{if } t_1 \leq t_2, \\
        \left({\mathbf{B}_{t_2\rightarrow t_1}}\right)^{-1} & \text{otherwise}  \end{cases}\label{equa:from_relative_to_absolution_pose} \;,
\end{align}
where $\mathbf{B}_{t_1\rightarrow t_2} =
    \begin{bmatrix}
        \mathbf{R}_{t_1\rightarrow t_2} & \mathbf{t}_{t_1\rightarrow t_2} \\
        \mathbf{0} & 1
    \end{bmatrix}\in\mathbb{R}^{4\times 4}$.
    
\subsection{Time-dependent NeRF}
\label{subsec:time_dependent_nerf}
Given the transformation matrices, an intuitive solution, as done in prior works~\cite{lin2021barf, bian2023nope}, is to train a single NeRF defined in the world coordinate system and map any 3D point to the world via these transformations. However, at early training stage, these transformations are very noisy which makes the optimization harder and often stuck in local minimum. To address this, we propose to relax the model from maintaining a global scene to many local ones by using a time-dependent NeRF to represent the scene at each time step. We then encourage the scene consistency across different time steps with soft constraints (i.e., losses). In particular, our NeRF model predicts the SDF $s(\mathbf{x},t)$ and the color $\mathbf{c}(\mathbf{x},t)$ for point $\mathbf{x}$ in the local camera coordinate system at time $t$ as,
\begin{equation}
    s(\mathbf{x},t) = \phi_g(\mathbf{x},t), \quad \mathbf{c}(\mathbf{x},t) = \phi_c({\mathbf{x}},\mathbf{d},t)\;,
\end{equation}
where $\mathbf{d}\in \mathbb{R}^{2}$ is a viewing direction in the camera coordinate system. We then perform volume rendering using Eq.~\ref{eq:volume_rendering} to obtain the color $\mathbf{\hat{C}}_t(\mathbf{r})\in\mathbb{R}^{3}$ for ray $\mathbf{r}$ at time $t$. 
The model is trained using the standard color loss and the Eikonal term \cite{gropp2020implicit} to constrain the SDF,
\begin{align}
    \mathcal{L}_{\text{rgb}} &= \frac{1}{|\Omega|}\sum_{r\in \Omega}{\left|\left|\mathbf{\hat{C}}_t(\mathbf{r})-\mathbf{C}_t(\mathbf{r})\right|\right|_2}\;,\\
    \mathcal{L}_{\text{eik}} &= \frac{1}{|\mathcal{X}|}\sum_{\mathbf{x} \in \mathcal{X}} \left( \left|\left| \frac{\partial s(\mathbf{x}, t)}{\partial \mathbf{x}} \right|\right|_2 - 1 \right)^2\;,
\end{align}
where $\Omega$ and $\mathcal{X}$ define a set of rays and sampled 3D points, respectively. In addition, we propose to use a series of losses, including $\mathcal{L}_{\text{flow}},\; \mathcal{L}_{\text{photo}},\; \mathcal{L}_{\text{sdf}} $ (see definition below), to encourage the consistency between the scene geometry across time, and the camera motion. 
Since the time-dependent NeRF models local scene at each time step, without constraint, there are inconsistencies in geometry for overlapping regions. To avoid this, we first leverage the linear relationship between scene flow and SDF flow derived in~\cite{maoneural} to constrain the camera motion and the scene geometry as,
\begin{equation}
    \mathcal{L}_{\text{flow}}=\frac{1}{|\mathcal{S}|}\sum_{\mathbf{x} \in \mathcal{S}} \left|\frac{\partial s(\mathbf{x},t)}{\partial t} + (\boldsymbol{\omega}(t) \times\mathbf{x} + \mathbf{v}(t))^T\mathbf{n}(\mathbf{x},t) \right|\;,
    \label{eq:SDF_flow_loss}
\end{equation}
where $\mathbf{n}(\mathbf{x})=\frac{\partial s(\mathbf{x},t)}{\partial \mathbf{x}}\in\mathbb{R}^3$ is the surface normal at $\mathbf{x}$ and $\mathcal{S}$ is the set of surface points. Intuitively, this loss constrains the scene change at different surface points to be consistent with the same camera motion at any time step, thus encouraging the scene surface to change rigidly with respect to time. For more details, please refer to the original paper~\cite{maoneural}.

We also utilize the photometric consistency loss from the literature on monocular depth estimation to supervise the camera motion and the time-dependent NeRF. To this end, we first obtain the surface point $\Tilde{\mathbf{x}}(\mathbf{r},t)\in \mathbb{R}^3$ at time step $t$ of ray $\mathbf{r}$ using volume rendering as,
\begin{equation}
    \Tilde{\mathbf{x}}(\mathbf{r},t) = \sum_{i=1}^{K} L_i \sigma_i \mathbf{x}_i(\mathbf{r},t)\;.
\end{equation}

We can then project the surface point to neighboring frames using the estimated camera pose,
\begin{equation}
    \Tilde{\mathbf{x}}'_{t\rightarrow t+n}(\mathbf{r},t) = \mathbf{P}_{t\rightarrow t+n} \Tilde{\mathbf{x}}'(\mathbf{r},t)\;,\label{eq:3d_correspond}
\end{equation}
where $n\in\mathcal{N}$ and $\mathcal{N}$ is a hyper-parameter defining a set of intervals to neighboring frames. Here $\Tilde{\mathbf{x}}'_{t\rightarrow t+n}(\mathbf{r},t)$ and $\Tilde{\mathbf{x}}'(\mathbf{r},t)$ denote the homogeneous coordinate of the corresponding points, $\Tilde{\mathbf{x}}_{t\rightarrow t+n}(\mathbf{r},t)$ and $\Tilde{\mathbf{x}}(\mathbf{r},t)$, respectively. Given these corresponding points, the photometric consistency loss is computed as,
\begin{equation}
\mathcal{L}_{\text{photo}} = \frac{1}{|\mathcal{N}| |\Omega|}\sum_{n,\mathbf{r}}{||\mathbf{C}_{t+n}[\mathbf{K} \Tilde{\mathbf{x}}_{t\rightarrow t+n}] - \mathbf{C}_t(\mathbf{r})||_1},
\end{equation}
where $\mathbf{K}$ defines the intrinsic matrix and $[\cdot]$ denotes the sampling operation. 

Lastly, we impose an SDF consistency loss $\mathcal{L}_{\text{sdf}}$ to encourage the consistency between the SDF values predicted in any frame and those in the world frame. In practice, we choose the middle frame as the world frame. 
\begin{equation}
    \mathcal{L}_{\text{sdf}} = \frac{1}{|\mathcal{X}|} \sum_{\mathbf{x} \in \mathcal{X}}  |s(\mathbf{x}, t)-s(\mathbf{x}_{{t}\rightarrow{t_w}}, t_w)|\;,
\end{equation}
where $t_w$ is the time step corresponding to the world frame, and $\mathbf{x}_{t\rightarrow t_w}$ is the corresponding point of $\mathbf{x}$ at the world frame computed via Eq.~\ref{eq:3d_correspond}.

\begin{table*}
\centering
\footnotesize
\begin{tblr}{
  colsep  = 3.20pt,
  rowsep  = 1pt,
  column{2} = {c},
  column{3} = {c},
  column{4} = {c},
  column{5} = {c},
  column{7} = {c},
  column{8} = {c},
  column{9} = {c},
  column{11} = {c},
  column{12} = {c},
  column{13} = {c},
  column{15} = {c},
  column{16} = {c},
  column{17} = {c},
  cell{1}{1} = {r=2}{},
  cell{1}{2} = {r=2}{},
  cell{1}{3} = {c=3}{},
  cell{1}{7} = {c=3}{},
  cell{1}{11} = {c=3}{},
  cell{1}{15} = {c=3}{},
  cell{1}{19} = {c=3}{c},
  cell{2}{6} = {c},
  cell{2}{10} = {c},
  cell{2}{14} = {c},
  cell{3}{1} = {r=4}{},
  cell{3}{14} = {c},
  cell{3}{19} = {c},
  cell{3}{20} = {c},
  cell{3}{21} = {c},
  cell{4}{14} = {c},
  cell{4}{19} = {c},
  cell{4}{20} = {c},
  cell{4}{21} = {c},
  cell{5}{14} = {c},
  cell{5}{19} = {c},
  cell{5}{20} = {c},
  cell{5}{21} = {c},
  cell{6}{14} = {c},
  cell{6}{19} = {c},
  cell{6}{20} = {c},
  cell{6}{21} = {c},
  cell{7}{1} = {r=5}{},
  cell{7}{19} = {c},
  cell{7}{20} = {c},
  cell{7}{21} = {c},
  cell{8}{19} = {c},
  cell{8}{20} = {c},
  cell{8}{21} = {c},
  cell{9}{19} = {c},
  cell{9}{20} = {c},
  cell{9}{21} = {c},
  cell{10}{19} = {c},
  cell{10}{20} = {c},
  cell{10}{21} = {c},
  cell{11}{19} = {c},
  cell{11}{20} = {c},
  cell{11}{21} = {c},
  hline{1,3,7,12} = {-}{},
  hline{2} = {3-5,7-9,11-13,15-17,19-21}{},
}
                                      & Scenes     & Ours                     &                          &                    &  & NeRFmm$^{\dagger\dagger}$ \cite{wang2021nerf} &                &            &  & NoPe-NeRF$^{*\dagger}$ (D) \cite{bian2023nope} &                &               &  & CF3DGS$^{\dagger\dagger}$(D) \cite{fu2023colmapfree} &                &               &  & CF3DGS$^{**}$(D) \cite{fu2023colmapfree}  &                &            \\
                                      &            & $\text{AbRel}\downarrow$ & $\text{SqRel}\downarrow$ & $\delta_1\uparrow$ &  & $\text{AbRel}$            & $\text{SqRel}$ & $\delta_1$ &  & $\text{AbRel}$             & $\text{SqRel}$ & $\delta_1$    &  & $\text{AbRel}$               & $\text{SqRel}$ & $\delta_1$    &  & $\text{AbRel}$   & $\text{SqRel}$ & $\delta_1$ \\
\begin{sideways}Scannet\end{sideways} & 0079       & \textbf{0.033}           & \textbf{0.004}           & \textbf{0.995}     &  & 0.184                     & 0.113          & 0.682      &  & 0.099                      & 0.047          & 0.904         &  & \underline{0.078}                & \underline{0.018}  & \underline{0.973} &  & --               & --             & --         \\
                                      & 0418       & \textbf{0.144}           & \textbf{0.067}           & \textbf{0.853}     &  & 0.611                     & 1.046          & 0.195      &  & \underline{0.152}              & \underline{0.137}          & \underline{0.738} &  & 0.239                        & {0.164}  & 0.593         &  & --               & --             & --         \\
                                      & 0301       & \textbf{0.037}           & \textbf{0.012}           & \textbf{0.977}     &  & 0.223                     & 0.208          & 0.562      &  & \underline{0.185}              & 0.252  & \underline{0.792} &  & 0.201                        & \underline{0.229}          & 0.745         &  & --               & --             & --         \\
                                      & 0431       & \textbf{0.038}           & \textbf{0.016}           & \textbf{0.982}     &  & 0.258                     & 0.187          & 0.503      &  & 0.127                      & 0.111          & 0.877         &  & \underline{0.108}                & \underline{0.052}  & \underline{0.900} &  & --               & --             & --         \\
\begin{sideways}Co3D\end{sideways}    & Bench      & \textbf{0.023}           & \textbf{0.085}           & \textbf{0.981}     &  & 0.291                     & 1.866          & 0.467      &  & \underline{0.183}              & \underline{0.945}  & 0.692         &  & 0.208                        & 3.028          & \underline{0.738} &  & --               & --             & --         \\
                                      & Skateboard & \textbf{0.012}           & \textbf{0.013}           & \textbf{0.998}     &  & 0.179                     & 0.539          & 0.674      &  & 0.159                      & \underline{0.527}  & 0.768         &  & \underline{0.138}                & 1.134          & \underline{0.865} &  & --               & --             & --         \\
                                      & Plant      & \textbf{0.074}           & \textbf{0.526}           & \textbf{0.933}     &  & 0.337                     & 4.145          & 0.444      &  & 0.249                      & \underline{1.816}  & 0.538         &  & \underline{0.201}                & 1.831          & \underline{0.650} &  & --               & --             & --         \\
                                      & Hydrant    & \textbf{0.022}           & \textbf{0.064}           & \textbf{0.988}     &  & 0.402                     & 4.281          & 0.260      &  & \underline{0.154}              & \underline{0.677}  & \underline{0.764} &  & 0.206                        & 1.950          & 0.651         &  & --               & --             & --         \\
                                      & Teddy      & \textbf{0.026}           & \textbf{0.175}           & \textbf{0.977}     &  & 0.254                     & 1.790          & 0.474      &  & \underline{0.135}              & \underline{0.823}  & \underline{0.842} &  & 0.302                        & 10.08          & 0.754         &  & --               & --             & --         
\end{tblr}
\caption{\textbf{Depth evaluation.} Our method achieves the best results across both datasets. (D) indicates methods leveraging depth priors. For each method, we report its released results if available (*). Otherwise, we train it using published implementation ($\dagger$). For example, NoPe-NeRF$^{*\dagger}$ means we report its released results on Scannet and train it using author's code for the Co3D dataset.
}
\label{tab:result_depth}
\end{table*}
\begin{table*}
\centering
\footnotesize
\begin{tblr}{
  colsep  = 2.65pt,
  rowsep  = 1pt,
  column{2} = {c},
  column{3} = {c},
  column{4} = {c},
  column{5} = {c},
  column{7} = {c},
  column{8} = {c},
  column{9} = {c},
  column{11} = {c},
  column{12} = {c},
  column{13} = {c},
  column{15} = {c},
  column{16} = {c},
  column{17} = {c},
  column{19} = {c},
  column{20} = {c},
  cell{1}{1} = {r=2}{},
  cell{1}{2} = {r=2}{},
  cell{1}{3} = {c=3}{},
  cell{1}{7} = {c=3}{},
  cell{1}{11} = {c=3}{},
  cell{1}{15} = {c=3}{},
  cell{1}{19} = {c=3}{},
  cell{3}{1} = {r=4}{},
  cell{3}{21} = {c},
  cell{4}{21} = {c},
  cell{5}{21} = {c},
  cell{6}{21} = {c},
  cell{7}{1} = {r=5}{},
  cell{7}{21} = {c},
  cell{8}{21} = {c},
  cell{9}{21} = {c},
  cell{10}{21} = {c},
  cell{11}{21} = {c},
  hline{1,7,12} = {-}{},
  hline{2} = {3-5,7-9,11-13,15-17,19-21}{},
  hline{3} = {3-21}{},
}
                                      & Scenes     & Ours                  &                       &                          &  & NeRFmm$^{\dagger\dagger}$ \cite{wang2021nerf} &               &                &  & NoPe-NeRF$^{*\dagger}$ (D) \cite{bian2023nope}&               &                &  & CF3DGS$^{\dagger\dagger}$(D) \cite{fu2023colmapfree} &               &                &  & CF3DGS$^{**}$ (D) \cite{fu2023colmapfree}  &               &                \\
                                      &            & $\text{PSNR}\uparrow$ & $\text{SSIM}\uparrow$ & $\text{LPIPS}\downarrow$ &  & $\text{PSNR}$           & $\text{SSIM}$ & $\text{LPIPS}$ &  & $\text{PSNR}$            & $\text{SSIM}$ & $\text{LPIPS}$ &  & $\text{PSNR}$               & $\text{SSIM}$ & $\text{LPIPS}$ &  & $\text{PSNR}$         & $\text{SSIM}$ & $\text{LPIPS}$ \\
\begin{sideways}Scannet\end{sideways} & 0079       & \textbf{35.52}        & \textbf{0.92}         & \uline{0.20}             &  & 32.06                   & \uline{0.85}  & 0.33           &  & 32.47                    & 0.84          & 0.41           &  & 33.91                       & \textbf{0.92}          & \textbf{0.16}           &  & --                     & --             & --              \\
                                      & 0418       & \textbf{34.53}        & \textbf{0.92}         & \uline{0.29}             &  & 30.92                   & 0.78          & 0.30           &  & 31.33                    & 0.79          & 0.34           &  & 32.56                       & 0.90          & \textbf{0.20}           &  & --                     & --             & --              \\
                                      & 0301       & \textbf{32.06}        & \textbf{0.88}         & \uline{0.22}             &  & 30.10                   & \uline{0.83}  & 0.29           &  & 29.83                    & 0.77          & 0.36           &  & 31.09                       & \textbf{0.88}          & \textbf{0.20}           &  & --                     & --             & --              \\
                                      & 0431       & \textbf{34.25}        & \textbf{0.93}         & \uline{0.25}             &  & 32.94                   & 0.90          & 0.31           &  & \uline{33.83}            & \uline{0.91}  & 0.39           &  & 31.02                       & 0.90          & \textbf{0.22}           &  & --                     & --             & --              \\
\begin{sideways}Co3D\end{sideways}    & Bench      & \textbf{26.36}        & 0.68                  & 0.40                     &  & 23.31                   & 0.58          & 0.55           &  & 24.32                    & 0.60          & 0.52           &  & 25.06                       & \uline{0.70}  & \uline{0.33}   &  & 26.21                 & \textbf{0.73} & \textbf{0.32}  \\
                                      & Skateboard & \textbf{30.93}        & \textbf{0.87}         & \uline{0.32}             &  & 23.10                   & 0.75          & 0.47           &  & 26.22                    & 0.80          & 0.42           &  & 22.21                       & 0.76          & 0.36           &  & 27.24                 & 0.85          & \textbf{0.30}  \\
                                      & Plant      & \uline{26.53}         & \uline{0.77}          & \uline{0.41}             &  & 22.26                   & 0.68          & 0.52           &  & 23.79                    & 0.71          & 0.48           &  & 20.73                       & 0.67          & 0.42           &  & \textbf{29.69}        & \textbf{0.89} & \textbf{0.29}  \\
                                      & Hydrant    & \uline{20.60}         & \uline{0.51}          & 0.44                     &  & 19.11                   & 0.41          & 0.59           &  & 19.82                    & 0.42          & 0.57           &  & 19.22                       & 0.50          & \underline{0.40}  &  & \textbf{22.14}        & \textbf{0.64} & \textbf{0.34}  \\
                                      & Teddy      & \textbf{33.04}        & \textbf{0.89}         & \textbf{0.20}            &  & 27.87                   & 0.77          & 0.42           &  & \uline{29.40}            & 0.80          & 0.38           &  & 24.13                       & 0.77          & 0.27           &  & 27.75                 & \uline{0.86}  & \uline{0.20}   
\end{tblr}
\caption{\textbf{Image evaluation.} Our approach significantly outperforms prior NeRF-based methods and is on par with CF3DGS. (D) indicates methods leveraging depth priors. For each method, we report its released results if available (*). Otherwise, we train it using published implementation ($\dagger$).}
\label{tab:result_image}
\end{table*}

\subsection{Training}
\label{subsec:overall_training_pipeline}
The overall loss used to train our model is,
\begin{equation}
    \mathcal{L} = 
    \mathcal{L}_{\text{rgb}} +
    \lambda_{1}\mathcal{L}_{\text{eik}} + 
    \lambda_{2}\mathcal{L}_{\text{flow}} + 
    \lambda_{3} \mathcal{L}_{\text{photo}} +
    \lambda_{4}\mathcal{L}_{\text{sdf}},
\end{equation}
where $\{\lambda_i\}_{i=1}^4$ are weights. Note that since the inferred camera-to-world transformations are initially noisy, we set the SDF consistency loss weight ($\lambda_{4}$) to be 0 for the first 200 epochs, and gradually increase it after that.

At the early stage of our training, we jointly train all models ($\phi_{v}$,\;$\phi_{g}$,\;$\phi_{c}$) to first obtain accurate camera motions, and the scene geometry encoded in different time steps. For the later stage of our training, we fix $t$ to the world time step $t_w$ and proceed with the conventional NeRF training pipeline \cite{mildenhall2021nerf} while keeping the camera poses $\mathbf{P}_{t\rightarrow t_w}$ fixed and only updating the scene geometry and appearance network ($\phi_g$, $\phi_c$) using all training frames in the video sequence. At this stage, we drop the loss weights $\lambda_2, \lambda_3, \lambda_4$ to 0. 

After the training, the full scene geometry and appearance are obtained at the world frame. It is also worth noting that our training pipeline is end-to-end trainable and does not rely on any prior knowledge about the camera motion or the scene.

\section{Experiments}

\begin{figure*}
    \centering
    \begin{subfigure}[b]{0.195\textwidth}
        \centering
        \caption*{Ground-truth}
        \includegraphics[width=\textwidth]{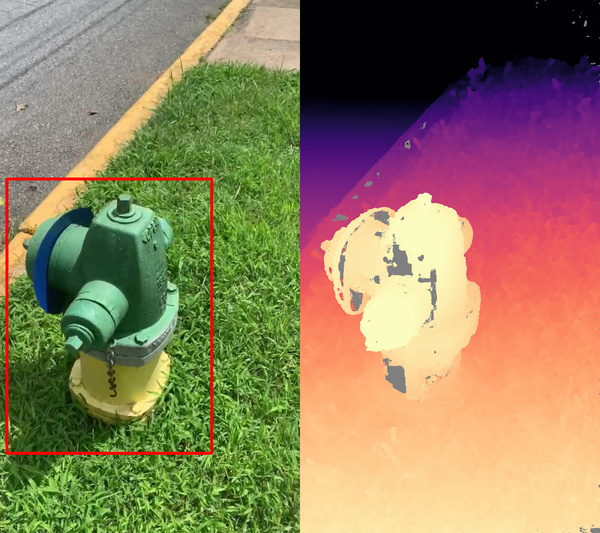}
    \end{subfigure}
    \begin{subfigure}[b]{0.195\textwidth}
        \centering
        \caption*{Ours}
        \includegraphics[width=\textwidth]{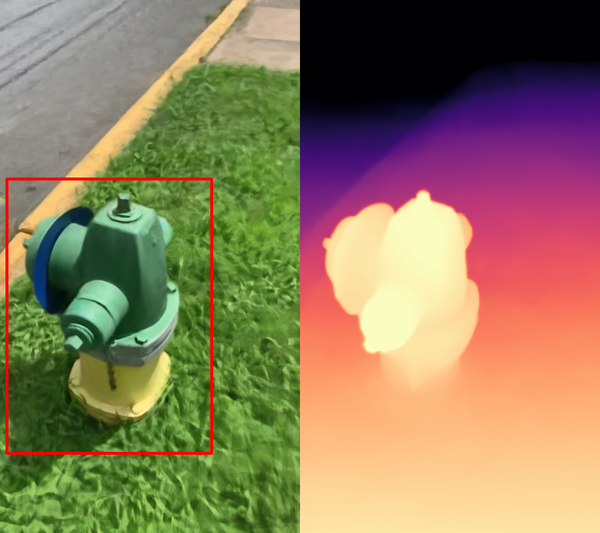}
    \end{subfigure}
    \begin{subfigure}[b]{0.195\textwidth}
        \centering
        \caption*{NeRFmm \cite{wang2021nerf}}
        \includegraphics[width=\textwidth]{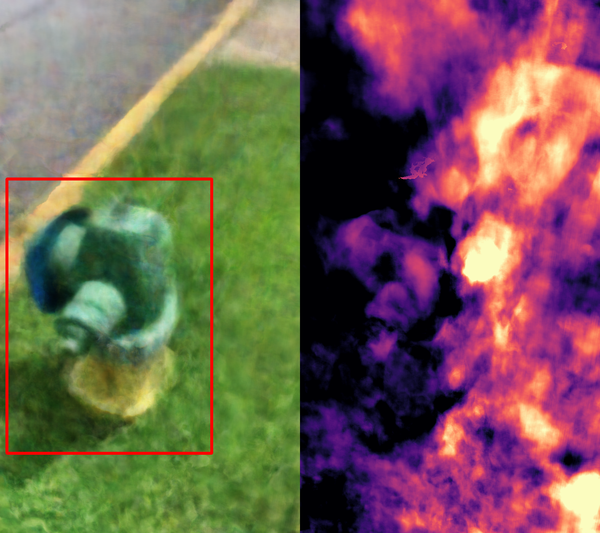}
    \end{subfigure}
    \begin{subfigure}[b]{0.195\textwidth}
        \centering
        \caption*{NoPe-NeRF \cite{bian2023nope}}
        \includegraphics[width=\textwidth]{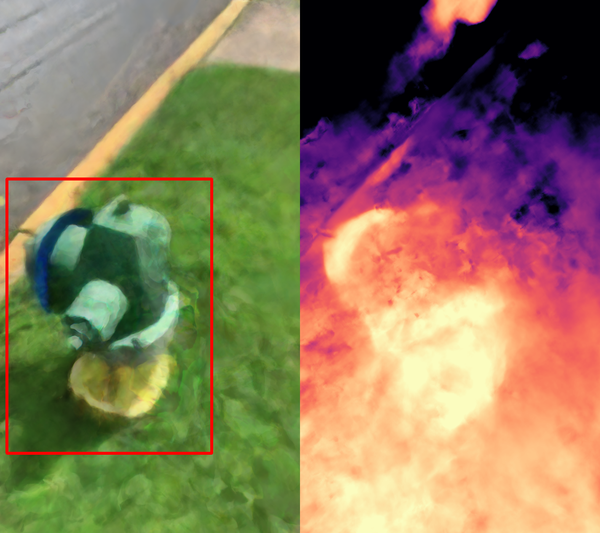}
    \end{subfigure}
    \begin{subfigure}[b]{0.195\textwidth}
        \centering
        \caption*{CF3DGS \cite{fu2023colmapfree}}
        \includegraphics[width=\textwidth]{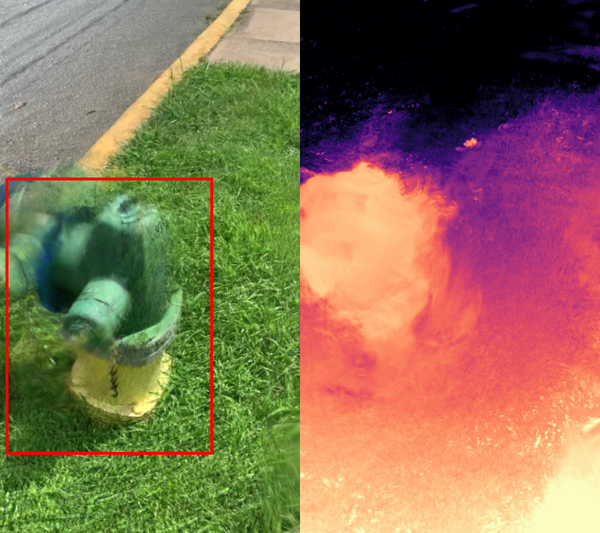}
    \end{subfigure}

    \begin{subfigure}[b]{0.195\textwidth}
        \centering
        \includegraphics[width=\textwidth]{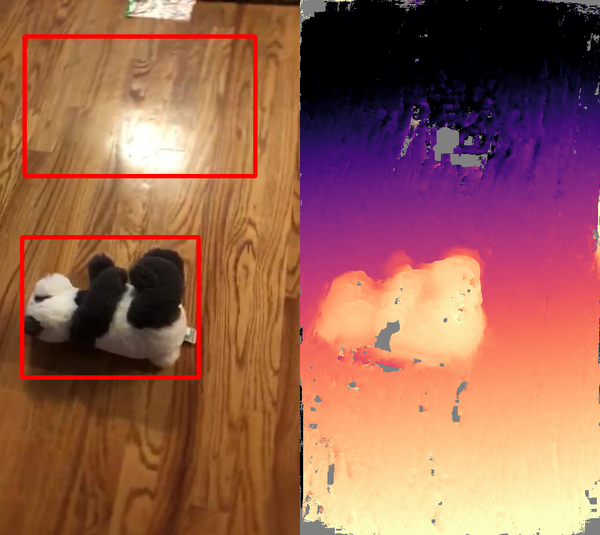}
    \end{subfigure}
    \begin{subfigure}[b]{0.195\textwidth}
        \centering
        \includegraphics[width=\textwidth]{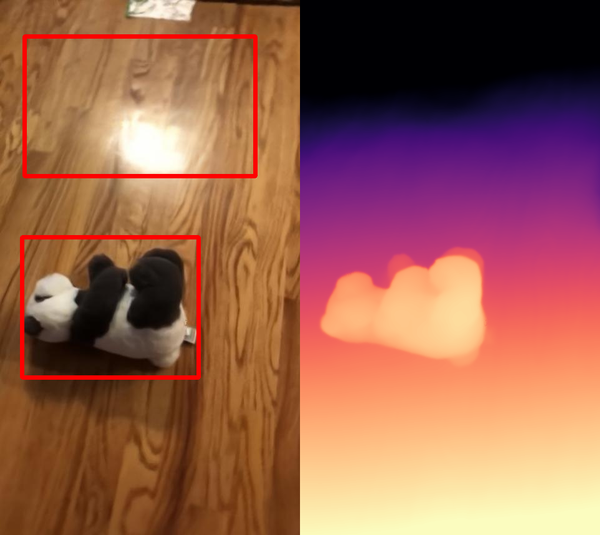}
    \end{subfigure}
    \begin{subfigure}[b]{0.195\textwidth}
        \centering
        \includegraphics[width=\textwidth]{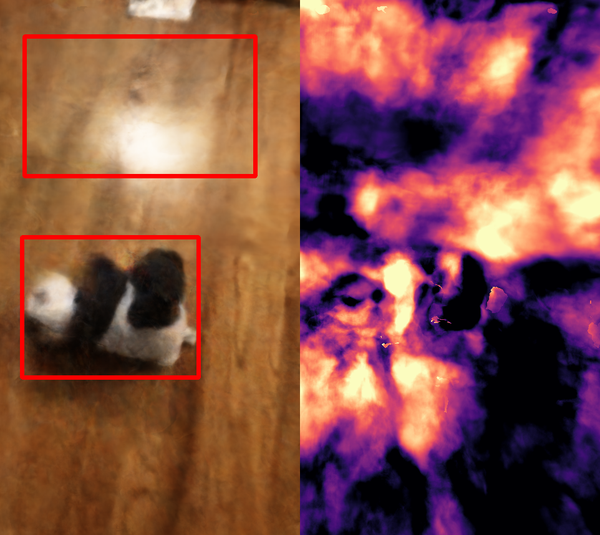}
    \end{subfigure}
    \begin{subfigure}[b]{0.195\textwidth}
        \centering
        \includegraphics[width=\textwidth]{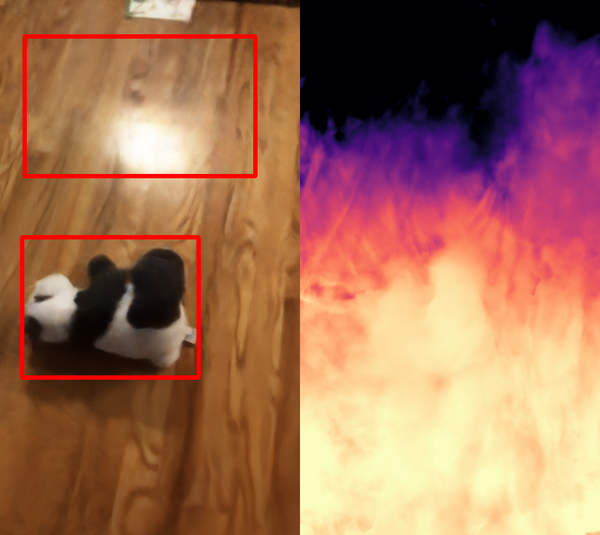}
    \end{subfigure}
    \begin{subfigure}[b]{0.195\textwidth}
        \centering
        \includegraphics[width=\textwidth]{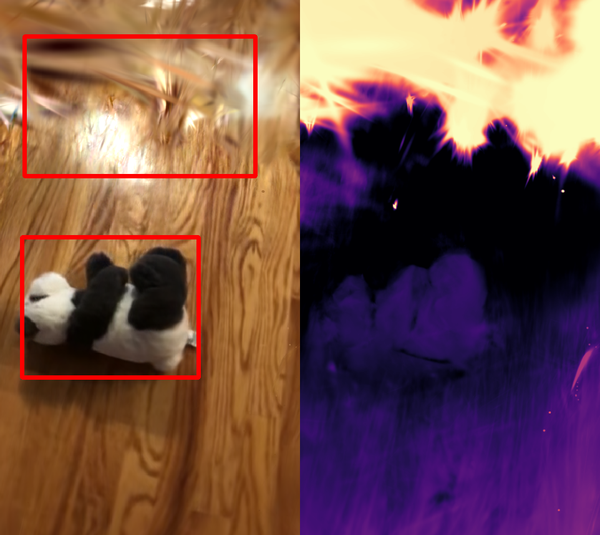}
    \end{subfigure}
    
    \vspace{-2.5mm}
    \hdashrule{\textwidth}{0.1pt}{2mm 1mm}
    \vspace{-3.0mm}

    \begin{subfigure}[b]{0.195\textwidth}
        \centering
        \includegraphics[width=\textwidth]{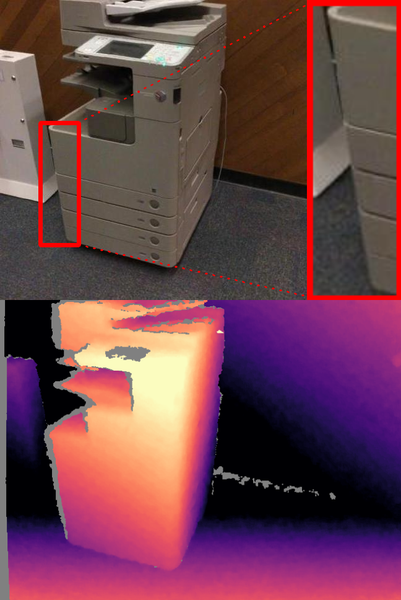}
    \end{subfigure}
    \begin{subfigure}[b]{0.195\textwidth}
        \centering
        \includegraphics[width=\textwidth]{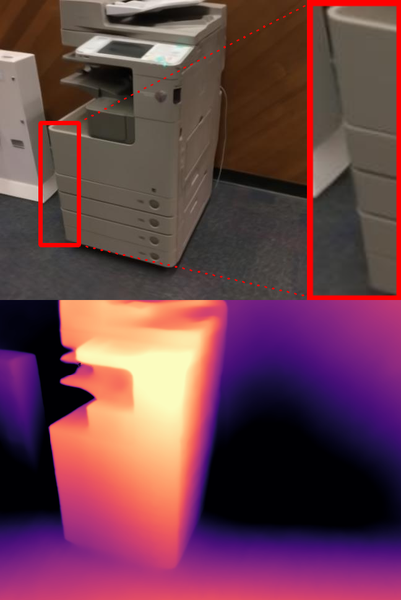}
    \end{subfigure}
    \begin{subfigure}[b]{0.195\textwidth}
        \centering
        \includegraphics[width=\textwidth]{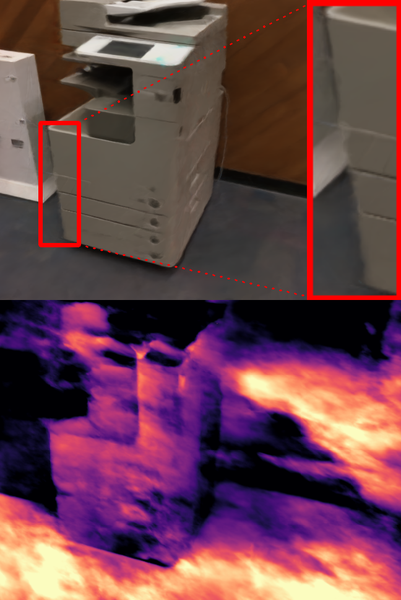}
    \end{subfigure}
    \begin{subfigure}[b]{0.195\textwidth}
        \centering
        \includegraphics[width=\textwidth]{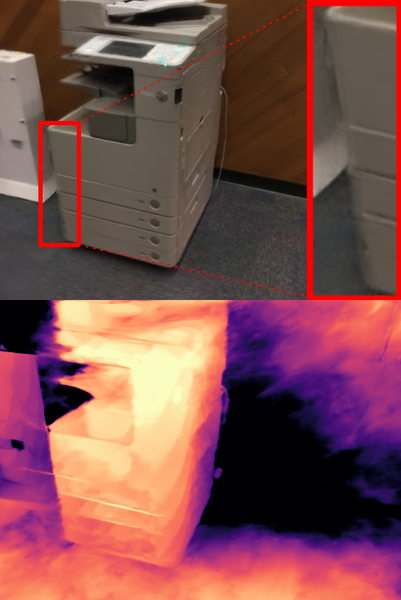}
    \end{subfigure}
    \begin{subfigure}[b]{0.195\textwidth}
        \centering
        \includegraphics[width=\textwidth]{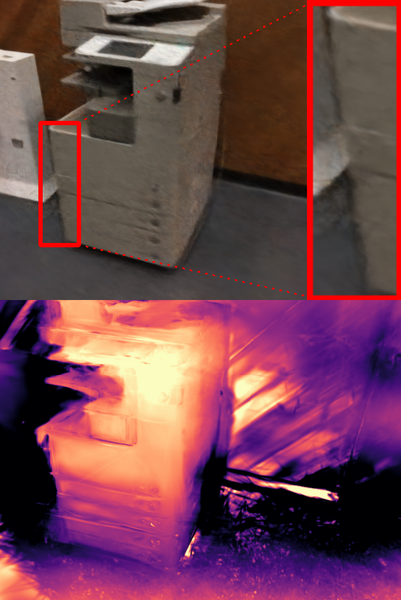}
    \end{subfigure}

    \caption{\textbf{Qualitative results on the Co3D  (top) and Scannet (bottom) dataset.}
    Our synthesized images are more photo-realistic compared to the other methods. 
    In terms of geometry, our method produces the most accurate depth maps among all methods.}
    \label{fig:qualitative_result}
\end{figure*}

\begin{figure*}
    \centering
    \begin{subfigure}[b]{0.24\textwidth}
        \includegraphics[width=\textwidth]{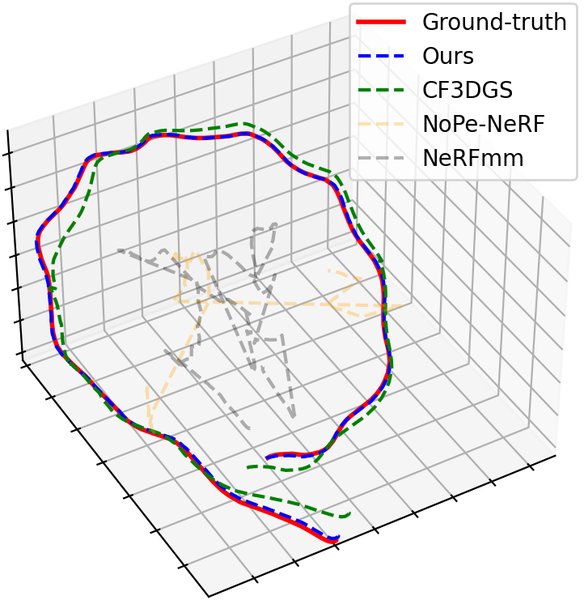}
        \caption*{Hydrant (Co3D)}
    \end{subfigure}
    \hfill
    \begin{subfigure}[b]{0.24\textwidth}
        \includegraphics[width=\textwidth]{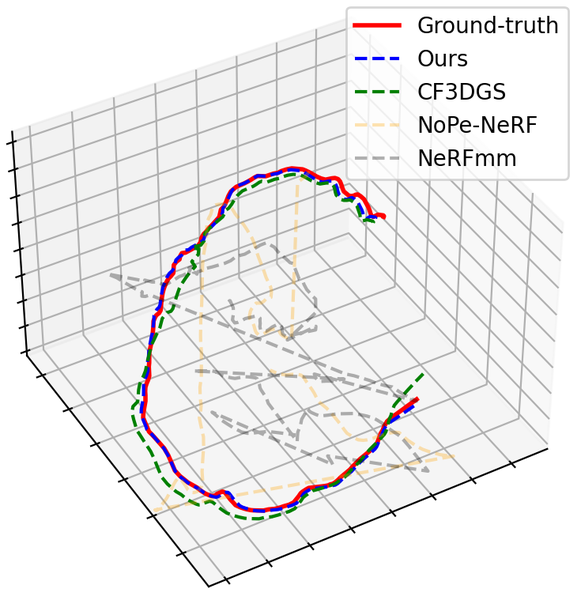}
        \caption*{Plant (Co3D)}
    \end{subfigure}
    \hfill
        \begin{subfigure}[b]{0.24\textwidth}
        \includegraphics[width=\textwidth]{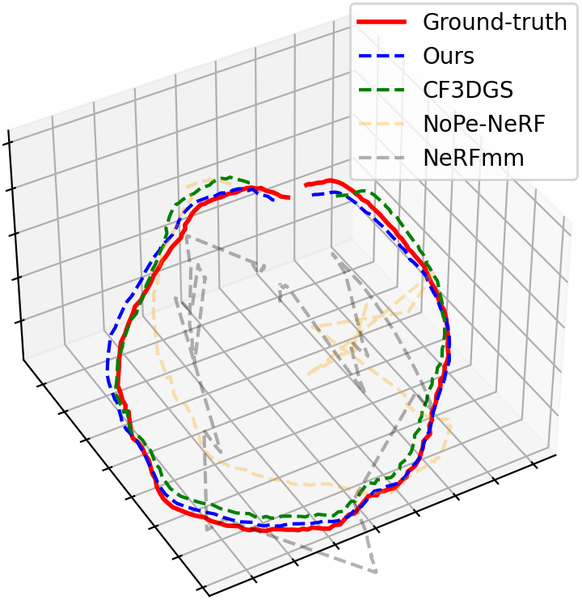}
        \caption*{Teddybear (Co3D)}
    \end{subfigure}
    \hfill
    \begin{subfigure}[b]{0.24\textwidth}
        \includegraphics[width=\textwidth]{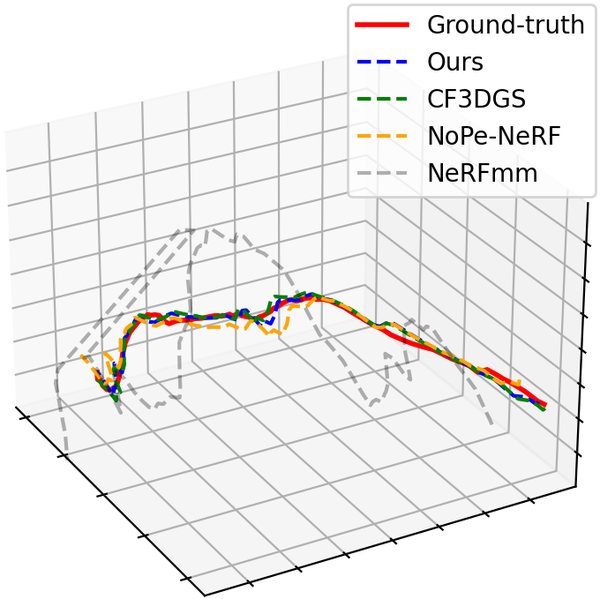}
        \caption*{0079 (Scannet)}
    \end{subfigure}
    
    \caption{\textbf{Camera trajectory visualization.} Our poses are better aligned with the ground-truth compared to the others.}
    \label{fig:pose_visualization}
\end{figure*}

\subsection{Experimental Setup}

\noindent\textbf{Dataset.}
Following \cite{bian2023nope, fu2023colmapfree}, we evaluate our method using 4 scenes from ScanNet~\cite{dai2017scannet} dataset and 5 scenes from Co3D ~\cite{reizenstein2021common} dataset. Each scene corresponds to a video containing a varying number of images. Similar to prior works, every 8\textit{th} image in each video is selected for novel-view synthesis and depth evaluation, whereas the rest are used for training. For both datasets, we utilize the provided poses and depth maps as ground-truth. While Scannet is captured in indoor environment, Co3D features a combination of indoor and outdoor scenes with more complex camera motions (i.e.,~large rotations). 
Moreover, we include results for the Tanks and Temples~\cite{knapitsch2017tanks} dataset in the supplementary materials.

\noindent\textbf{Metrics.} We follow a similar evaluation scheme as ~\cite{bian2023nope}. Specifically, the metrics used for novel-view synthesis are PSNR, SSIM~\cite{wang2004image} and LPIPS~\cite{zhang2018unreasonable}. Regarding pose evaluation, we use RPE$_t$, RPE$_r$ and ATE to measure the errors of the relative translations, relative rotations, and camera trajectories, respectively. Prior to computing these metrics, we apply Umeyama alignment \cite{umeyama1991least} to align the estimated and ground-truth poses. For depth evaluation, we report three metrics including the absolute relative error (AbRel), square relative error (SqRel) and a depth accuracy metric ($\delta_1$). Since the rendered depths are up-to-scale, we follow \cite{zhou2017unsupervised} to recover the ground-truth scale before calculating the depth metrics.   

\noindent\textbf{Implementation Details.} Our NeRF architecture is based on \cite{wang2021neus} with two modifications: (1) We use the SDF-network to represent the entire scene instead of employing an extra model \cite{mildenhall2021nerf} for the background, and (2) our model takes a timestep as an additional input. 
The motion network has a similar structure to the SDF-network, with the exception of the input and output layers. 
The input timesteps for all models are normalized from $-1$ to $1$. We use Adam \cite{kingma2017adammethodstochasticoptimization} optimizer with a learning rate of $0.001$ to train the time-dependent NeRF and the motion network until convergence. Then, we fine-tune our NeRF model following the conventional training pipeline \cite{mildenhall2021nerf} for an additional 5000 epochs with the learning rate decaying gradually. At each training iteration, we sample 1024 rays and 128 3D points per ray within a pre-defined depth range. For all scenes, we use 10 sub-intervals between two consecutive frames to solve for the relative camera transformation between them (Eq.~\ref{eq:relative_rot} and~\ref{eq:relative_trans}), and the world time step $t_w$ is chosen as the middle frame in the training set. More details regarding the datasets, evaluation metrics and other hyper-parameters can be found in the supplementary materials.

\begin{table*}
\centering
\footnotesize
\begin{tblr}{
  colsep  = 3.25pt,
  rowsep  = 1pt,
  column{2} = {c},
  column{3} = {c},
  column{4} = {c},
  column{5} = {c},
  column{7} = {c},
  column{8} = {c},
  column{9} = {c},
  column{11} = {c},
  column{12} = {c},
  column{13} = {c},
  column{15} = {c},
  column{16} = {c},
  column{19} = {c},
  column{20} = {c},
  column{21} = {c},
  cell{1}{1} = {r=2}{},
  cell{1}{2} = {r=2}{},
  cell{1}{3} = {c=3}{},
  cell{1}{7} = {c=3}{},
  cell{1}{11} = {c=3}{},
  cell{1}{15} = {c=3}{},
  cell{1}{19} = {c=3}{},
  cell{2}{14} = {c},
  cell{3}{1} = {r=4}{},
  cell{3}{14} = {c},
  cell{3}{17} = {c},
  cell{4}{14} = {c},
  cell{4}{17} = {c},
  cell{5}{14} = {c},
  cell{5}{17} = {c},
  cell{6}{14} = {c},
  cell{6}{17} = {c},
  cell{7}{1} = {r=5}{},
  cell{7}{17} = {c},
  cell{8}{17} = {c},
  cell{9}{17} = {c},
  cell{10}{17} = {c},
  cell{11}{17} = {c},
  hline{1,3,7,12} = {-}{},
  hline{2} = {3-5,7-9,11-13,15-17,19-21}{},
}
                                      & Scenes     & Ours                     &                          &                        &  & NeRFmm$^{\dagger\dagger}$ \cite{wang2021nerf} &                &              &  & NoPe-NeRF$^{*\dagger}$ (D) \cite{bian2023nope} &                &                &  & CF3DGS$^{\dagger\dagger}$ (D) \cite{fu2023colmapfree} &                &               &  & CF3DGS$^{**}$ (D) \cite{fu2023colmapfree}  &                &               \\
                                      &            & $\text{RPE}_t\downarrow$ & $\text{RPE}_r\downarrow$ & $\text{ATE}\downarrow$ &  & $\text{RPE}_t$          & $\text{RPE}_r$ & $\text{ATE}$ &  & $\text{RPE}_t$           & $\text{RPE}_r$ & $\text{ATE}$   &  & $\text{RPE}_t$              & $\text{RPE}_r$ & $\text{ATE}$  &  & $\text{RPE}_t$  & $\text{RPE}_r$ & $\text{ATE}$  \\
\begin{sideways}Scannet\end{sideways} & 0079       & \textbf{0.664}           & \textbf{0.182}           & \textbf{0.013}         &  & 3.312                   & 0.801          & 0.151        &  & 0.752                    & 0.204          & 0.023          &  & \uline{0.673}               & \uline{0.190}  & \uline{0.015} &  & --              & --             & --            \\
                                      & 0418       & \textbf{0.401}           & \textbf{0.119}           & \textbf{0.015}         &  & 1.428                   & 0.396          & 0.016        &  & \uline{0.455}            & \textbf{0.119} & \textbf{0.015} &  & 0.529                       & 0.131          & 0.019         &  & --              & --             & --            \\
                                      & 0301       & \textbf{0.367}           & \textbf{0.117}           & \textbf{0.009}         &  & 1.852                   & 0.591          & 0.296        &  & 0.399                    & 0.123          & \uline{0.013}  &  & \uline{0.393}               & \uline{0.118}  & 0.015         &  & --              & --             & --            \\
                                      & 0431       & \textbf{1.097}           & \textbf{0.223}           & \textbf{0.042}         &  & 6.661                   & 0.574          & 0.475        &  & 1.625                    & 0.274          & 0.069          &  & \uline{1.301}               & \uline{0.269}  & \uline{0.064} &  & --              & --             & --            \\
\begin{sideways}Co3D\end{sideways}    & Bench      & \textbf{0.013}           & \textbf{0.037}           & \textbf{0.001}         &  & 0.562                   & 3.508          & 0.039        &  & 0.301                    & 1.925          & 0.053          &  & \uline{0.059}               & \uline{0.357}  & 0.017         &  & 0.110           & 0.424          & \uline{0.014} \\
                                      & Skateboard & \textbf{0.024}           & \textbf{0.063}           & \textbf{0.001}         &  & 0.702                   & 2.794          & 0.057        &  & 0.421                    & 1.883          & 0.048          &  & \uline{0.063}               & \uline{0.462}  & \uline{0.017} &  & 0.239           & 0.472          & \uline{0.017} \\
                                      & Plant      & \textbf{0.014}           & \textbf{0.062}           & \textbf{0.002}         &  & 0.650                   & 2.557          & 0.059        &  & 0.305                    & 1.587          & 0.047          &  & \uline{0.049}               & 0.465          & \uline{0.006} &  & 0.140           & \uline{0.401}  & 0.021         \\
                                      & Hydrant    & \textbf{0.010}           & \textbf{0.027}           & \textbf{0.001}         &  & 0.497                   & 2.556          & 0.063        &  & 0.337                    & 1.557          & 0.060          &  & \uline{0.059}               & \uline{0.344}  & \uline{0.008} &  & 0.094           & 0.360          & \uline{0.008} \\
                                      & Teddy      & \textbf{0.053}           & \textbf{0.129}           & \textbf{0.004}         &  & 0.588                   & 3.025          & 0.049        &  & 0.286                    & 1.295          & 0.040          &  & \uline{0.072}               & \uline{0.184}  & \uline{0.009} &  & 0.505           & 0.211          & \uline{0.009} 
\end{tblr}
\caption{\textbf{Pose evaluation}. Our method achieves the lowest pose errors in all scenes. (D) indicates methods leveraging depth priors. For each method, we report its released results if available (*). Otherwise, we train it using published implementation ($\dagger$).}
\label{tab:result_pose}
\end{table*}

\begin{table}
\centering
\scriptsize
\begin{tblr}{
  colsep  = 1.5pt,
  rowsep  = 0.75pt,
  column{2} = {c},
  column{3} = {c},
  column{5} = {c},
  column{6} = {c},
  column{7} = {c},
  column{9} = {c},
  column{10} = {c},
  cell{1}{2} = {c=2}{},
  cell{1}{5} = {c=3}{},
  cell{1}{9} = {c=2}{},
  hline{1,3,10} = {-}{},
  hline{2} = {2-3,5-7,9-10}{},
}
               & Depth  &            &  & Pose  &       &       &  & Image  &       \\
               & $\text{AbRel}\downarrow$ & $\delta_1\uparrow$ &  & $\text{RPE}_t\downarrow$  & $\text{RPE}_r\downarrow$  & $\text{ATE}\downarrow$   &  & $\text{PSNR}\uparrow$   & $\text{SSIM}\uparrow$  \\
Full           & \textbf{0.031}  & \textbf{0.975}      &  & \textbf{0.023} & \textbf{0.063} & \textbf{0.002} &  & \underline{27.49} & \underline{0.74} \\
w/o~$\mathcal{L}_{\text{flow}}$ & 0.084  & 0.877      &  & 0.114 & 0.371 & 0.008 &  & 26.31 & 0.72 \\
w/o~$\mathcal{L}_{\text{sdf}}$ & \underline{0.072}  & \underline{0.913}      &  & 0.031 & \underline{0.094} & \underline{0.004} &  & 26.54 & 0.72 \\

w/o~$\mathcal{L}_{\text{photo}}$      & 0.399  & 0.386      &  & 0.216 &	1.402 &	0.041     &  & 21.79  &	0.65 \\
w/o $\text{NeRF}_{t}$      & 0.287  & 0.504      &  & 0.299     & 1.808     & 0.045     &  & 24.08 & 0.67 \\
w/o $\text{NeRF}_{t\rightarrow{c}}$      & 0.082  & 0.857      &  & --     & --     & --     &  & \textbf{28.29} & \textbf{0.76} \\
w/o motion net      & 0.080  & 0.912      &  & \underline{0.028}	& 0.154	& 0.006     &  & 26.23	& 0.71 \\ 
\end{tblr}
\caption{Ablation study on the Co3D dataset. }
\vspace{-4mm}
\label{tab:ablation_study}
\end{table}

\subsection{Results}

We compare our method with two other NeRF-based methods: NeRFmm~\cite{wang2021nerf} and NoPe-NeRF~\cite{bian2023nope}, and a recent 3DGS-based method:  CF3DGS~\cite{fu2023colmapfree}. Among these, our method and NeRFmm take only RGB images from a monocular video as input, whereas NoPe-NeRF and CF3DGS leverage additional geometric information from a pre-trained depth network.

\noindent\textbf{Geometry. } We quantify the geometric accuracy of each method through depth evaluation in Tab. \ref{tab:result_depth}. Our method consistently outperforms other methods with notable margins across all scenes in both datasets. On average, our method reduces the depth square relative errors by at least {\textbf{82\%}} and {\textbf{78\%}} on the Co3D and Scannet datasets, respectively. Fig. \ref{fig:qualitative_result} reveals that although both NoPe-NeRF and CF3DGS leverage depth priors, they still fail to render accurate depth maps. On the other hand, our method can produce correct scene geometry given only RGB images during training.  

\noindent\textbf{Novel-view Synthesis. } To obtain the camera poses corresponding to the test images, we keep the NeRF model fixed and learn these poses by minimizing the color rendering loss $\mathcal{L}_{\text{rgb}}$, as in \cite{bian2023nope}. The results in Tab. \ref{tab:result_image} show that our method significantly outperforms prior NeRF-based methods and is comparable to CF3DGS. However, as can be seen in Fig.  \ref{fig:intro_figure} and Fig.  \ref{fig:qualitative_result}, CF3DGS produces clear artifacts in all of its rendered images on the Co3D dataset. On the other hand, our synthesized images are visually more photorealistic than those of CF3DGS. 

\noindent\textbf{Camera Poses. } Tab. \ref{tab:result_pose} highlights the superior performance of our method in recovering the camera poses. While there are minor improvements on Scannet dataset, our method demonstrates a clear dominance in all challenging scenes in the Co3D dataset. Specifically, on Co3D, we achieve a minimum of \textbf{83\%}, \textbf{63\%} and \textbf{83\%} error reductions in estimating the camera trajectory, relative camera translation and relative camera rotation, respectively. 
The inferior performance of NeRFmm and NoPe-NeRF is primarily due to their direct optimization of the camera-to-world transformations, which is challenging in case of large camera motions. In contrast, our method can avoid this issue and correctly recover the camera poses by modeling continuous camera motions. It can be observed from Fig. \ref{fig:pose_visualization} that our learned camera trajectories are better aligned with the ground truth compared to the other methods. Additionally, we also provide a comparison with COLMAP in the supplementary materials, demonstrating that our method yields more accurate camera poses.

\subsection{Ablation Study}
In this section, we conduct an ablation study to demonstrate the effectiveness of each component in our proposed method. The results are shown in Tab. \ref{tab:ablation_study}.

\textbf{w/o $\boldsymbol{\mathcal{L}_{\text{flow}}}$}: Without $\mathcal{L}_{\text{flow}}$, there is a significant performance drop in all aspects. \textbf{w/o $\boldsymbol{\mathcal{L}_{\text{sdf}}}$}: Training without the SDF consistency loss $\mathcal{L}_{\text{sdf}}$ leads to performance degradation. \textbf{w/o $\boldsymbol{\mathcal{L}_{\text{photo}}}$:} Dropping the photometric loss results in the largest errors in all metrics. This illustrates the effectiveness of the losses used in our method.

\textbf{w/o $\text{NeRF}_{\boldsymbol{t}}$}: Instead of leveraging a time-dependent NeRF, we aggregate the predicted motions to obtain the camera poses, and use them to train a conventional NeRF \cite{wang2021neus}. As the aggregated motions are initially noisy, this strategy results in the large errors across all metrics. \textbf{w/o $\text{NeRF}_{\boldsymbol{t\rightarrow{c}}}$}: Instead of fine-tuning the time-dependent NeRF in the later training stage, we use our learned camera poses to train a new conventional NeRF model from scratch. This approach slightly improves the quality of the rendered images but causes an increase of roughly 160\% in the depth error. 
\textbf{w/o motion net:} Here, we remove the motion network and optimize the camera poses as learnable variables. Consequently, the error increases significantly. These results highlight the significance of the time-dependent NeRF and the continuous motion modeling in our pipeline.

\vspace{-0.3cm}
\section{Conclusion}
In this work, we present a prior-free pipeline for the joint optimization of camera poses and NeRF from a monocular video. The key contributions are to model continuous camera motions and adopt a time-dependent NeRF
to avoid the challenge of directly optimizing large camera-to-world mappings and obtain accurate geometry.
Extensive experiments on Co3D and Scannet dataset demonstrate that our method outperforms previous methods in terms of pose estimation and depth estimation, while achieving comparable performance in novel-view synthesis.

\noindent\textbf{Limitation.} Since our method relies on visual cues in the training images to learn the camera motions, the joint optimization can become challenging in scenes having large low-texture regions or reflective surfaces. 

\noindent\textbf{Acknowledgement.}
This research was supported in part by the Australia Research Council ARC Discovery Grant (DP200102274)).

{
    \small
    \bibliographystyle{ieeenat_fullname}
    \bibliography{main}
}


\end{document}